\documentclass{article}

\usepackage[final]{neurips_2020}

\usepackage[utf8]{inputenc} 
\usepackage[T1]{fontenc}    
\usepackage{hyperref}       
\usepackage{url}            
\usepackage{booktabs}       
\usepackage{amsfonts}       
\usepackage{nicefrac}       
\usepackage{microtype}      
\usepackage{amsmath}
\usepackage{amsthm}
\usepackage{amssymb}
\usepackage{dsfont}
\usepackage{mathtools}
\usepackage{enumerate}
\usepackage{cleveref}
\usepackage{caption}
\usepackage{subcaption}
\usepackage{graphicx}
\usepackage{float}
\usepackage{multirow}
\usepackage{booktabs}
\usepackage{multicol}

\newtheorem{corollary}{Corollary}[section]
\newtheorem{definition}{Definition}[section]

\newtheorem{lemma}{Lemma}[section]

\newtheorem{theorem}{Theorem}[section]

\DeclareMathOperator*{\esssup}{ess\,sup}

\newcommand{\cvar}{\textnormal{CVaR}_{\alpha}}

\usepackage[dvipsnames]{xcolor}
\definecolor{MyDarkBlue}{rgb}{0,0.08,0.45}
\def\boxit#1{\vbox{\hrule\hbox{\vrule\kern6pt
          \vbox{\kern6pt#1\kern6pt}\kern6pt\vrule}\hrule}}

\title{Risk-Averse Action Selection Using Extreme
  Value Theory Estimates of the CVaR}

\author{%
Dylan Troop\thanks{Corresponding author.}\\
  Department of Computer Science and Software Engineering\\
  Concordia University\\
  Montreal, Canada \\
  \texttt{d\_troop@encs.concordia.ca} \\
  \And
  Fr\'ed\'eric  Godin \thanks{Quantact Actuarial and Financial Mathematics
  Laboratory, Montreal, Canada.} \\
  Department of Mathematics and Statistics \\
  Concordia University\\
  Montreal, Canada \\
  \texttt{frederic.godin@concordia.ca} \\
  \And
  Jia Yuan Yu \\
  Concordia Institute of Information Systems Engineering \\
  Concordia University \\
  Montreal, Canada \\
  \texttt{jiayuan.yu@concordia.ca} \\
}

\begin{document}

\maketitle

\begin{abstract}
In a wide variety of sequential decision making problems, it can be important to estimate the impact of rare events in order to minimize risk exposure. A popular risk measure is the conditional value-at-risk (CVaR), which is commonly estimated by averaging observations that occur beyond a quantile at a given confidence level. When this confidence level is very high, this estimation method can 
exhibit high variance due to the limited number of samples above the corresponding quantile. To mitigate this problem, extreme value theory can be used to derive an estimator for the CVaR that uses extrapolation beyond available samples. This estimator requires the selection of a threshold parameter to work well, which is a difficult challenge that has been widely studied in the extreme value theory literature. In this paper, we present an estimation procedure for the CVaR that combines extreme value theory and a recently introduced method of automated threshold selection by \cite{bader2018automated}. Under appropriate conditions, we estimate the tail risk using a generalized Pareto distribution. We compare empirically this estimation procedure with the commonly used method of sample averaging, and show an improvement in performance for some distributions. We finally show how the estimation procedure can be used in reinforcement learning by applying our method to the multi-arm bandit problem where the goal is to avoid catastrophic risk.
\end{abstract}
\section{Introduction}
In the stochastic multi-arm bandit (MAB) problem, a learning agent is
presented with the repeated task of selecting from a number of
choices (arms), each providing independent and identically distributed
rewards. The agent has no prior knowledge of the reward
distributions. Through feedback observation of the reward with a combination of
exploration and exploitation, the agent attempts to identify the arm with the most favorable reward distribution; see
\cite{lattimore_2020} for a description of such a setting.

In the traditional MAB framework, the most favorable distribution maximizes the expected reward over
time. However, more recent generalizations of this problem have been
considered in the literature where the expectation objective is
replaced by other metrics aimed at measuring risk. For instance,
\cite{sani2012risk,yu2013sample,pmlr-v29-Galichet13,David2018PACBW,torossian2019xarmed,NIPS2019_9347,ravi_tails,NIPS2019_9305} address the MAB problem with a risk-averse agent. The risk considered may
either be instantaneous, i.e., considering risk for a single draw of a
reward, or cumulative, i.e., considering jointly all
subsequent rewards.

The agent may be interested in minimizing the impact of a rare
catastrophic loss. Risk measures targeted at quantifying exposure to
extreme losses are well studied in the risk management literature. A
popular example introduced by \cite{rockafellar2002conditional} is the
\textit{conditional value-at-risk} (CVaR), which measures the average loss given that
the latter exceeds a given quantile of its distribution. Theoretical
properties of the CVaR risk measure are studied in, for example,
\cite{acerbi2002coherence} and \cite{sarykalin2008value}. Note that not all risk measures are exclusively targeting catastrophic risk; other measures also quantify the impact of moderate unfavorable outcomes, see for instance the semi-variance. Nevertheless, the objective of the current paper is to tackle extreme risk minimization, which makes CVaR a suitable choice in this context.

An important challenge that the agent faces when using the CVaR as the
objective function in the MAB context is the estimation
of the CVaR from a finite sample of observations. If an extreme
quantile confidence level is given for the CVaR, the sparsity of
observations lying in the tail of the distribution can yield imprecise
results in common calculation methods such as sample averaging.  We propose to employ results from \textit{extreme value theory} (EVT) to obtain better estimates
of the CVaR, a method which has been discussed in, for example, \citet[Section 7.2.3]{mcneil2005quantitative}. In particular, the Pickands-Balkema-de Haan theorem \citep{pickands1975statistical,balkema1974residual}
presents a parametric approximation of the tail data using a
\textit{generalized Pareto distribution} (GPD). The theorem states that by selecting an
appropriate threshold, the distribution of tail data beyond that
threshold can be well-approximated by the GPD. The parametric modeling
of the tail distribution is often referred to the
\textit{peaks-over-threshold} (POT) approach, which is investigated in, for example, \cite{simiu1996extreme, ferreira1998application, frigessi2002dynamic, begueria2006mapping, GKILLAS2018109}.

The major drawback of the POT approach is that it can be difficult to select a threshold that fits the GPD model well. The threshold selection problem presents a bias-variance tradeoff: too high a threshold results in limited data availability causing high variance, whereas too low a threshold can cause a large bias between the GPD and true tail distribution. Therefore, estimating the CVaR using the POT approach can also prove challenging. In this paper, we are motivated by recent advancement in threshold selection algorithms to investigate CVaR estimation using the POT approach. We combine the POT methodology with
the recent work of \cite{bader2018automated} for automated threshold selection via ordered goodness-of-fit tests to estimate the CVaR. This methodology has recently shown promising experimental results for estimating high quantiles, i.e., \cite{extreme_tail_risk_pot}. Using this estimation procedure, an application to a risk-averse MAB problem is then presented. While EVT has been used to estimate reward distributions in the MAB setting, i.e., \cite{NIPS2014_evtbandits}, this work is, to the best of our knowledge, the first to use EVT in the MAB setting under risk criteria.

The current paper is divided as follows. In \cref{sec:ProbFormul}, the
notation used in the current work is introduced and the risk-averse
MAB problem is defined along with the CVaR risk measure. In
\cref{se:TCEEVTsec}, a background on EVT is
provided, and an approach to estimate the CVaR using EVT is
illustrated. In \cref{sec:CVaRapprox}, the statistical estimation
procedures used for the CVaR calculation is discussed, including the
automated threshold selection procedure of \cite{bader2018automated}. In
\cref{sec:MultiArmBandits}, details of the MAB policy in
a risk-averse setting are discussed. In \cref{sec:SimulStudies},
results from numerical simulations comparing statistical estimation
procedures for the CVaR in the multi-arm bandit setting are
shown. Some proofs are provided in \cref{se:Proofs}.
 
\section{Problem Formulation}
\label{sec:ProbFormul}

\subsection{The multi-arm bandit framework}

The MAB framework involves a finite horizon
multi-stage decision setting, where an agent makes decisions at stages
$t=1,\ldots,T$.  Let $\mathbb{K} \equiv \{1,...,k\}$ denote a set of
arms, which are possible actions that can be taken at each stage. In the risk-averse setting, we consider the outcome of each draw from an arm to be  a cost to the
agent (i.e., the larger the value that is sampled, the more unfavorable the outcome is considered).  For $t=1,\ldots,n$, define
the $k$-dimensional random vector $X^t \equiv (X_1^t,\ldots,X_k^t)$
where $X^t_j$ denotes the cost incurred if the arm $j$ is selected at
stage $t$. Vectors $X^1, \ldots, X^n$ are assumed to be independent
and identically distributed. Therefore, for all arms $i=1,\ldots,k$,
cost variables $X^1_i,\ldots,X^n_i$ are i.i.d. $\!\!\!$ copies of some
random variable $X_i$.  Let $\{F_1, \ldots, F_k\}$ denote the
respective cumulative distribution functions (CDF) of
$X_1,\ldots,X_k$; these distribution functions are unknown to the
agent.

The sequence of selected arms is denoted by
$a \equiv (a_1,\ldots,a_n)$ where $a_t$ is the random variable taking
values in $1,\ldots,k$ denoting the arm selected at time $t$. When an
arm $a_t$ is selected at time $t$, its associated cost $X^t_{a_t}$ is
observed, but the costs associated with all other arms
$\{X^t_i : i\neq a_t \}$ remain unobserved.

The selection of one of the $k$ arms at each time step is decided
through a \textit{policy}. A policy is a mapping
that returns the probabilities of selecting any action at the next stage given the agent's current state. The policy evolves over time as new samples are obtained and
results in a sequence of policies $\pi_1,\ldots,\pi_n$ where, for a
given $t$, the function
$\pi_t : \prod_{i=1}^{t-1}(\mathbb{K} \times \mathbb{R} ) \rightarrow [0,1]^k$
takes as input all previous realizations of
actions and costs,
$\left((a_{1}, X^1_{a_{1}}), \ldots, (a_{t-1}, X^{t-1}_{a_{t-1}})
\right)$, and maps them into probabilities of selecting any possible next-stage action $a_t$.

Policies considered in the current paper attempt to identify the arm
with the least risk, as quantified through a risk measure.  Let $\chi$
denote a set of random variables. For a given confidence level
$\alpha \in (0,1)$, let $\rho_{\alpha} : \chi \rightarrow \mathbb{R}$
denote a law-invariant\footnote{A measure $\rho$ is said to be law
  invariant if $X$ and $Y$ having the same distribution implies
  $\rho(X)=\rho(Y)$.} risk measure.

Since the cost probability distributions are a priori unknown, every
time an arm is sampled, the estimate of the risk associated with the
sampled arm is refined. The notation $\hat{\rho}_{\alpha}^t(X_i)$ is
used to refer to the estimate of $\rho_{\alpha}(X_i)$ after the first
$t$ stages.  The least risky arm is denoted
$i^* = \underset{i \in \mathbb{K}}{\arg \min} \,\rho_{\alpha}(X_{i})$.

\subsection{The CVaR risk measure}
In this section we define the CVaR and introduce its commonly used estimator.
For a given random variable $Y$, we denote the CVaR at a confidence level $\alpha$ as $\cvar(Y)$ along with its CDF $F_Y$, the quantile
of confidence level $\alpha$ of the distribution of $Y$ is defined as
\begin{equation*}
  q_{\alpha} = \inf\{x\in\mathbb{R} : F_Y(x) \geq \alpha\}.
\end{equation*}
This allows to define in turn the CVaR as in
\cite{rockafellar2002conditional} as the mean of the $\alpha$-tail
distribution, $F^{(\alpha)}_Y$ of $Y$, which has the following CDF:
\begin{eqnarray*}
  F^{(\alpha)}_Y (y) \equiv \begin{cases}
    0 \text{ if } y < q_{\alpha},
    \\ \frac{F_Y (y) -\alpha}{1-\alpha} \text{ if } y \geq q_{\alpha}.
  \end{cases}
\end{eqnarray*}
Typical values of $\alpha$ are $0.95$, $0.99$ or $0.999$.
If the random variable $Y$ is
absolutely continuous, it can be shown that
\begin{equation*}
  \cvar(Y) = \mathbb{E}[Y | Y \geq q_{\alpha}],
\end{equation*}
which gives and intuitive interpretation to the CVaR. Without loss of
generality, the current work will only consider absolutely continuous
variables for simplicity. Note that all results in the current work could be easily generalized to consider the optimization of a risk-reward tradeoff by selecting an
objective function of the form
$\rho_{\alpha}(Y) \equiv \mathbb{E}[Y] + \lambda \,\cvar(Y)$ instead of
the purely risk-centric framework $\rho_{\alpha} \equiv \cvar$.

\subsubsection{Sample average CVaR estimation}
\label{se:SampleTCE}

Since for each arm $j$ the CDF $F_{X_j}$ is unknown, it must be
estimated from costs previously sampled from the arm $j$. Consider an
i.i.d. sample $S_t = \{y_1, \ldots, y_t\}$ of observations drawn from
a distribution $F_Y$. For every $y \in \mathbb R$, the sample CDF
estimator (i.e., empirical distribution function) is defined as
\begin{equation}
  \label{SampleCDFestim}
  \hat{F}^t_Y(y) \equiv t^{-1} \sum_{s=1}^{t} \mathds{1}_{ \{ y_s \leq y \} }.
\end{equation}
The sample CDF can be plugged into the definition of the quantile and
the CVaR to obtain simple estimators of these quantities. Let
$\{y_{(1)}, \ldots, y_{(t)}\}$ be the set of order statistics, i.e.,
the observations sorted in non-decreasing order. Then, the empirical quantile estimator is
\begin{eqnarray}
  \label{quantileEstim}
  \hat{q}^t_{\alpha} \equiv \inf\{x\in\mathbb{R} : \hat{F}^t_Y(y) \geq \alpha\} = \underset{i }{\min}\{y_{(i)} : \hat{F}^t_Y(y_{(i)}) \geq \alpha\} = y_{(\lceil\alpha t\rceil)},
\end{eqnarray}
and in turn the \textit{sample average} CVaR
estimator is 
\begin{equation}
  \label{eq:TCEestimsample}
  \widehat{\cvar^t}(Y)  
  = \frac{\sum_{i=1}^{t} y_{i} \mathds{1}_{ \{y_i \geq \hat{q}^t_{\alpha} \} }}{\sum_{i=1}^{t} \mathds{1}_{ \{y_i \geq \hat{q}^t_{\alpha} \} }}.
\end{equation}


A confidence interval for the sample CVaR estimate can be obtained
through bootstrapping as described in \cref{se:STCECI}. 
Such confidence intervals can be useful to design lower-confidence-bound action selection schemes, which are a direct analogue of upper-confidence bound algorithms (see \citet[Chapter 7]{lattimore_2020}) in the risk-averse setting. Such schemes are left out-of-scope of the current paper.

\section{Estimating the CVaR through extreme value theory}
\label{se:TCEEVTsec}

The use of the sample CDF to estimate $\cvar$ can be problematic when
the sample size is small and the confidence level $\alpha$ is large. The
scarcity of sampled observations lying in the tail of the distribution
can lead to a volatile estimate of the tail distribution and thus of
the CVaR.
We therefore turn to extreme value theory, which was developed in an attempt to estimate
the tail distribution from scarce samples by exploiting
the asymptotic behavior of the tail distribution above increasingly
high quantiles. This section shows how to use extreme
value theory to approximate the CVaR, and in turn to estimate the approximation from
i.i.d. observations.

\subsection{The Pickands-Balkema-de Haan theorem and CVaR
  approximation}

For a random variable $Y$ with CDF $F_Y$ and a given threshold
$u > \text{ess inf} \, Y$, the \textit{excess distribution function} $K_{u}$
is defined for $z>0$ as
\begin{align*}
  K_{u}(z) &\equiv \mathbb{P}(Y-u \leq z | Y > u) \\
             &= \frac{\mathbb{P}(Y-u \leq z, Y > u)}{\mathbb{P}(Y > u)} \\
             &= \frac{\mathbb{P}(u < Y \leq z+u)}{\mathbb{P}(Y > u)} \\
             &= \frac{F_Y(z+u)-F_Y(u)}{1-F_Y(u)}.
\end{align*}
Note that the domain of $K_{u}$ is $[0,\esssup Y)$.  The $z$-values are
referred to as the \textit{threshold excesses}. Given that $Y$ has
exceeded some high threshold $u$, this function represents the
probability that it exceeds the threshold by at most $z$. When $F_Y$
is unknown, $K_{u}$ cannot be calculated directly, but can be approximated by the generalized Pareto distribution (GPD).

\begin{definition}[GPD]
  \label{def:GPD}
  The \textit{generalized Pareto distribution} (GPD) with two
  parameters $\xi \in \mathbb{R}$ and $\sigma>0$ is a continuous
  probability distribution with PDF
  \begin{equation}
    \label{GPD-PDF}
    g_{\xi, \sigma}(y) = \begin{cases}
      \frac{1}{\sigma}\left (1+\frac{\xi y}{\sigma}\right )^{(-1/\xi - 1)}, \quad 0 \leq y \leq -\sigma/\xi \quad \text{ if } \xi < 0,
      \\ \frac{1}{\sigma}\left (1+\frac{\xi y}{\sigma}\right )^{(-1/\xi - 1)}, \quad 0 \leq  y <\infty \quad \text{ if } \xi > 0,
      \\ \frac{1}{\sigma}\exp \left(-\frac{y}{\sigma}\right), \quad \quad \,\,\, \quad 0 \leq y <\infty \quad  \text{ if } \xi = 0,
      \\ 0 \quad\quad \text{otherwise}.
    \end{cases}
  \end{equation}
  Over its support, the CDF is given by
  \begin{equation}
    \label{def:GP}
    G_{\xi, \sigma}(y) = \begin{cases}
      1-\left (1+\frac{\xi y}{\sigma}\right )^{(-1/\xi)} \quad \textrm{ if } \xi \neq 0,
      \\ 1- \exp \left(-\frac{y}{\sigma}\right), \quad y \geq 0\quad \textrm{ if } \xi = 0.
    \end{cases}
  \end{equation}
\end{definition}

The Pickands-Balkema-de Haan theorem states that under certain conditions and for any large enough $u$, $K_{u}$ is well approximated by the GPD. Two additional definitions are needed to state the theorem.

\begin{definition}[GEV]
  The generalized extreme value (GEV) distribution with single
  parameter $\xi\in\mathbb R$ has CDF
  \begin{equation*}
    H_\xi(y) \equiv \begin{cases}
      \exp \left(-(1+\xi y )^{(-1/\xi)}\right)
      \quad \text{ if } \xi \neq 0,
      \\ \exp \left(-e^{-y}\right)\quad \text{ if } \xi = 0
    \end{cases}
  \end{equation*}
  over its support, which is $[-1/\xi,\infty)$ if $\xi > 0$,
  $(-\infty,-1/\xi]$ if $\xi < 0$ or $\mathbb{R}$ if $\xi=0$.
\end{definition}

\begin{definition}[MDA]
  Let $F$ denote the CDF of some random variable and let $H_\xi$
  denote the GEV with parameter $\xi$. $F$ is
  said to belong to the \textit{Maximum Domain of Attraction} (MDA) of $H_\xi$, which is denoted $F \in \text{MDA}(H_\xi)$, if
  there exist real sequences $\{c_n\}^\infty_{n=0}$ and
  $\{d_n\}^\infty_{n=0}$ with $c_n \geq 0$ such that
  \begin{equation*}
    \underset{n \rightarrow \infty}{\lim } F^n \left(c_n y + d_n\right) = H_\xi(y),
  \end{equation*}
  for all $y \in \mathbb{R}$.
\end{definition}

\begin{theorem}[Pickands-Balkema-de Haan]
  \label{PickhandsBalkema}
  Consider a real value $\xi$ and a random variable $Y$ such that
  $y_{\max} \equiv \esssup Y \leq \infty$ and that
  $F_Y \in \text{MDA}(H_\xi)$. Then there exists a positive function
  $\beta$ such that
  \begin{equation*}
    \underset{u \rightarrow y_{\max}}{\lim} \,\,\underset{0 \leq z \leq y_{\max} - u}{\sup} \vert K_{u}(z) - G_{\xi, \beta(u)}(z) \vert = 0.
  \end{equation*}
\end{theorem}

The property $F_Y \in \text{MDA}(H_\xi)$ for some $\xi$ holds for a
large class of distributions, in particular it holds for all common
continuous distributions (e.g., uniform, normal, Student,
exponential, beta, Fr\'echet, etc).

Using \cref{PickhandsBalkema}, an approximation for the CVaR can be derived. The following result can be found in, for example, \citet[Section 7.2.3]{mcneil2005quantitative}.

\begin{corollary}[CVaR Approximation]
  \label{cor:approxTCE}
  Consider a random variable $Y$ such that $F_Y \in \text{MDA}(H_\xi)$
  for some $\xi<1$.  Consider $u$ sufficiently large with
  $u \leq q_\alpha$, where $q_\alpha$ is the quantile of confidence level
  $\alpha$ of $Y$.
  Then,
  \begin{equation}
    \cvar(Y) \approx q_\alpha + \frac{\beta(u) + \xi(q_\alpha-u)}{1-\xi}.
    \label{eq:TCEapproxEVT}
  \end{equation}
  where $\beta$ is the function specified in \cref{PickhandsBalkema}.
\end{corollary}


\section{Statistical estimation of the CVaR approximation}
\label{sec:CVaRapprox}

In practice, using the CVaR approximation \eqref{eq:TCEapproxEVT}
requires identifying suitable values for the threshold $u$ and
parameters $\xi$ and $\sigma=\beta(u)$ from a sample of observations
$S_t=\{y_1,\ldots,y_t\}$. Such considerations are discussed in the
current section.

\subsection{Estimating \texorpdfstring{$(\xi,\sigma)$}{(Xi, Sigma)} for a given threshold}


First, assume that the threshold $u$ is pre-determined, and that
parameters $\xi$ and $\sigma$ are estimated based on such a choice
$u$. The maximum likelihood approach for the estimation of such
parameters is a typical procedure. Consider the set of excesses over
the threshold $u$ defined by
\begin{equation*}
  \mathcal{Z}_u \equiv \{ y_i - u \vert y_i \geq u, i=1,\ldots,t \}.
\end{equation*}
Elements of
$\mathcal{Z}_u$ 
are i.i.d \citep[Section 3.4]{dehaan2006} and approximately distributed
as GPD$(\xi,\sigma)$ with $\sigma= \beta(u)$ for some mapping $\beta$
by \cref{PickhandsBalkema}. The maximum likelihood estimator entails
solving the following optimization problem:
\begin{equation}
  \label{eq:MLestim}
  (\hat{\xi}, \hat{\sigma}) = \underset{\xi, \sigma}{\arg\max} \sum_{z \in \mathcal{Z}_u} \log g_{\xi, \sigma}(z),
\end{equation}
where $g_{\xi, \sigma}$ is defined in \eqref{GPD-PDF}. Such an
optimization must be conducted numerically as closed-form solutions to
this problem are not available. In the current paper, since we want to
consider integrable distributions (so that the CVaR exists), the constraint $\xi<1$ is imposed when the maximum likelihood
optimization \eqref{eq:MLestim} is applied.

This leads to an estimate of $\cvar(Y)$ of based on
\eqref{eq:TCEapproxEVT}:
\begin{equation}
  \widehat{\cvar}(Y) \approx \hat{q}_\alpha + \frac{\hat{\sigma} + \hat{\xi}(\hat{q}_\alpha-u)}{1-\hat{\xi}},
  \label{eq:TCEestimEVT}
\end{equation}
where $(\hat{\xi}, \hat{\sigma})$ are obtained from
\eqref{eq:MLestim}. An approximate asymptotic confidence interval for the CVaR
estimate can be derived by combining the asymptotic maximum likelihood
variance of parameter estimates and the delta method, see
\cref{se:EVTCI}. The misspecification of the tail distribution, i.e. the fact that the
conditional tail distribution is not exactly a GPD distribution in
general, causes the estimator \eqref{eq:TCEestimEVT} to be
asymptotically biased in general. The construction of the confidence interval
based on the delta method also disregards the conditional tail
distribution misspecification issue, which leads to a loss in
precision.

\subsection{Estimating the extreme quantile with EVT}
The calculation of $\cvar(Y)$ requires determining its quantile
$q_\alpha$. A first possibility would be to use the estimate
given by \eqref{quantileEstim}. However, extreme value theory can also
be used for such purpose.

Assume that the threshold $u$ that is used in the CVaR estimation
procedure is smaller than the quantile of interest, i.e.
$q_\alpha \geq u$. Denote $\hat{q}_\alpha$ as the estimate of
$q_\alpha$, and recall \eqref{SampleCDFestim} which defines
$\hat{F}^t_Y$ as the empirical CDF generated by
$S_t = \{y_1, \ldots, y_t\}$, a sample from i.i.d. copies of $Y$. The
following results gives the approximation formula for $q_\alpha$ which
relies on \cref{PickhandsBalkema}. Without loss
of generality, only the result for $\xi\neq0$ is provided, with a similar interpretation for $\xi=0$ based on \cref{def:GP}.
\begin{corollary}
  \label{co:VarApprox}
  Assume that $q_\alpha \geq u$ and that $F_Y \in \text{MDA}(H_\xi)$
  for some $\xi >0$. Then the quantile $q_\alpha$ of the distribution
  of $Y$ can be approximated through
  \begin{equation}
    \label{eqcorVaR}
    \hat{q}_\alpha=u+\frac{\hat{\sigma}}{\hat{\xi}}\left[\left(\frac{1-\alpha}{1-\hat{F}^t_Y(u)}\right)^{-\hat{\xi}_n}-1\right].
  \end{equation}
\end{corollary}

\subsection{Choosing the threshold} \label{se:threshold}
The selection of a suitable threshold $u$ is a much harder
problem that has been well-studied in the extreme value theory literature. For a survey of approaches for setting the threshold, see \cite{scarrott2012review}.  Many such approaches involve applying
judgment to ultimately select a value of $u$. Typically, sensitivity
analyses are performed by altering the threshold values and ensuring
results are robust to the choice of $u$. However, a challenging aspect
of threshold selection in the machine learning context of the current
paper is that $u$ must be decided automatically. We apply the recently
developed method of \cite{bader2018automated}, which uses a
combination of ordered goodness-of-fits tests and a stopping rule to
choose the optimal threshold
automatically. This method provides some assurance that the excesses above the chosen threshold are sufficiently well approximated the GPD.
The method of \cite{bader2018automated} is as follows. Consider a
fixed set of thresholds $u_1 < \ldots < u_l$, where for each $u_i$ we
have $n_i$ excesses.  The sequence of null hypotheses for each
respective test $i$, $i=1,\ldots,l$ is given by
$$H_0^{(i)}:\text{The distribution of the $n_i$ excesses above $u_i$ follows the GPD}.$$
For each threshold $u_i$, the Anderson-Darling (AD) test statistic
comparing the empirical threshold exceedances distribution and the GPD
is calculated. Let $z_1 < ... < z_{n_i}$ be the ordered threshold
exceedances for test $i$, and $\hat{\theta}_i$ the corresponding MLE
estimate of parameters for the GPD. The transformation
$\mathcal{U}^{(i)}_{(j)} \equiv G_{\hat{\theta}_i}(z_j)$ for
$1 < j < n_i$ is applied, where $G$ is the GPD CDF from
\eqref{def:GP}. The AD statistic is then
$$A_{i}^{2}=-n_i-\frac{1}{n_i} \sum_{j=1}^{n_i}(2 j-1)\left[\log \left(\mathcal{U}^{(i)}_{(j)}\right)+\log \left(1-\mathcal{U}^{(i)}_{(n_i+1-j)}\right)\right].$$
Corresponding $p$-values for each test statistic can then be found by
referring to a lookup table (e.g., \citet{stephens}) or computed on-the-fly.  Finally, using the
$p$-values $p_1, \ldots, p_l$ calculated for each test, the ForwardStop
rule of \cite{gsell} is used to choose the threshold. This is done by
calculating a cutoff
\begin{equation}
\label{BaderIndex}
\hat{k}_F=\max \left\{k \in\{1, \ldots, l\} :-\frac{1}{k} \sum_{i=1}^{k} \log \left(1-p_{i}\right) \leq \gamma\right\},
\end{equation}
where $\gamma$ is a chosen significance parameter. Under this rule,
the threshold $u_{\hat{k}_F+1}$ is chosen. If no $\hat{k}_F$ exists,
then no rejection is made and $u_1$ is chosen.

Thus, summarizing the overall tail distribution estimation procedure, the threshold and GPD parameter estimates are respectively provided by
\begin{eqnarray*}
u &\equiv& \begin{cases}
u_{\hat{k}_F+1} \, \text{ if the set in \eqref{BaderIndex} is not empty},
\\ u_1 \text{ otherwise},
\end{cases}
\\ (\hat{\xi}, \hat{\sigma}) &=& \underset{\xi, \sigma}{\arg\max} \sum_{z \in \mathcal{Z}_u} \log g_{\xi, \sigma}(z).
\end{eqnarray*} 

\section{Multi-arm bandit policies}
\label{sec:MultiArmBandits}

The current section outlines the proposed policies that are
investigated in the simulation study of the next section for the
context of multi-arm bandit (MAB) problems.

For each considered policy, after each stage $t$, an estimate
$\widehat{\cvar^t}(X_j)$ is available for all arms. Such estimates can
be used to determine the action at the subsequent stage. The CVaR
estimates for all arms allow defining an $\epsilon$-greedy policy
which is now described.  Consider the following deterministic sequence
$\epsilon \equiv \{\epsilon_t\}^n_{t=1}$ containing real numbers in
$[0,1]$. The sequence $\epsilon$ is referred to as a
\textit{schedule}. $\epsilon_t$ defines the probability of making an
exploratory move at stage $t$ instead of exploiting knowledge
(i.e. selecting the perceived least risky action). Typically, the
schedule is a decreasing sequence so as to progressively reduce the
amount of exploration as the cost distributions estimated become more
precise.  Let $\Pi_{t,j}$ be the probability of selecting action $j$
at stage $t$. Such quantities characterize the policy followed by the
agent.  The $\epsilon$-greedy policy entails choosing the action at
stage $t$ according to the following rule:
\begin{equation*}
  \Pi_{t,j}= \begin{cases}
    1-\epsilon_t+\epsilon_t/k \quad \text{ if } j =  \underset{i \in \mathbb{K}}{\arg \min} \,\widehat{\cvar^{t-1}}(X_i),
    \\ \epsilon_t/k \quad \text{otherwise}.
  \end{cases}
\end{equation*}
In other words, at stage $t$ such a policy entails choosing randomly and
uniformly across all arms with a probability $\epsilon_t$, or
selecting the greedy action (i.e. the one with the least estimated
risk) with probability $1-\epsilon_t$. When more than a single action
reaches the minimal estimated risk among all arms (i.e. when the $\arg \min$
set is not a singleton), the arm with the minimum index is selected to
break the tie.

To determine the estimates $\widehat{\cvar^t}(X_j)$, two methodologies
are compared.  The first estimation approach contemplated is the
sample CVaR estimation stemming from
\eqref{quantileEstim}-\eqref{eq:TCEestimsample}. This approach is
referred to subsequently as the \textit{Sample Average} (SA) method.
The second methodology considered involves the extreme value theory
estimator outlined in \cref{se:TCEEVTsec} and
\cref{sec:CVaRapprox}. The description of such an approach referred to
as the \textit{Extreme Value Theory} (EVT) method is provided next.

For each arm $j$, let $S^j_t = \{ y^j_s : a_s = j, s=1,\ldots,t \}$ be
the sample containing all costs sampled from arm $j$ between stage
$1$ and $t$. The number of elements of the set $S^j_t$ is denoted
$N^j_t \equiv \sum^t_{s=1} \mathds{1}_{ \{a_s = j\} }$. Before stage
$1$, all CVaR estimates are set to zero:
\begin{equation*}
  \widehat{\cvar^0}(X_j) \equiv 0.
\end{equation*}
Subsequently, each time an action $j$ is selected at some stage $t$,
the associated CVaR estimate is refined based on the new cost outcome
generated by arm $j$. To update the CVaR estimate, a threshold
$u^{t}_j$ is selected based on observations $S^j_t$. The set of
threshold exceedances over the threshold $u^{t}_j$ computed from the
set $S^j_t$ are then used to estimate the corresponding Generalized
Pareto distribution parameters as indicated in
\eqref{eq:MLestim}. This allows using \eqref{eq:TCEestimEVT} as the
updated CVaR estimate $\widehat{\cvar^t}(X_j)$, where the quantile
$q_\alpha$ is estimated according to \eqref{eqcorVaR}.  For all other
arms, i.e., for all $\ell \neq a_t$, the CVaR estimate is left untouched
i.e. $\widehat{\cvar^t}(X_\ell) \equiv \widehat{\cvar^{t-1}}(X_\ell)$.

Throughout the rest of the paper, it is assumed that the reward
distribution associated with each arm satisfies the MDA assumption,
i.e. that for all $j=1,\ldots,k$, there exists $\xi_j <1$ such that
$F_{X_j} \in \text{MDA}(H_{\xi_j})$. Such an assumption is not very
restrictive as it holds for a very large class of distributions. The
integrability assumption underlying $\xi_j <1$ is neither very
restrictive in practice. This implies that the estimate
\eqref{eq:TCEestimEVT} is valid to approximate the CVaR associated
with any arm $j$, i.e. $\cvar(X_j)$, provided the threshold $u$ is
sufficiently large for each arm.

\section{Simulation Studies}
\label{sec:SimulStudies}

In this section, both Sample Average (SA) and Extreme Value Theory
(EVT) CVaR estimation methods described in the previous section are
compared within a simulation study.  Two simulation experiments will
be conducted. The first is a pure statistical estimation problem where
i.i.d. costs from a single arm are sequentially observed, and the cost
distribution CVaR estimated based on both respective methods are
updated every time a new observation becomes available. This allows
evaluating the statistical accuracy of both methods. The second
simulation experiment embeds the two respective CVaR estimation
methods within a MAB problem so as to assess their
suitability for sequential action selection. The current section
provides details about these experiments and outlines numerical results obtained.\footnote{Code to replicate our results can be found in the following repository: \url{https://github.com/dtroop/evt-bandits}.}

\subsection{Single-arm CVaR estimation experiment}
\label{se:singlearm}

The single-arm problem where all costs are i.i.d. samples from an
unknown distribution is first considered. The estimation performance
of the SA and EVT methods is compared. The experiments consist of
performing $M=1,\!000$ independent runs. Each run consists in
sequentially sampling $n=5000$ independent costs from the single arm,
and every time a new sample is observed the CVaR estimates are updated
according to both respective methods.

Three families of distributions are considered for the arm costs: GPD,
Weibull (WE) and lognormal (LN). The density of the last two is given
by
\begin{eqnarray*}
  f^{(WE)}(x;\kappa, \lambda) &=& \frac{\kappa}{\lambda} \left(\frac{x}{\lambda}\right)^{k-1} \exp \left(- \left(\frac{x}{\lambda} \right)^k\right), \quad x>0,
  \\ f^{(LN)}(x;\mu,\sigma) &=& \frac{1}{\sqrt{2\pi}\sigma x} \exp \left( -\frac{\left(\log x-\mu \right)^2}{2\sigma^2}  \right), \quad x>0.
\end{eqnarray*}

Such distributions are chosen since the exact value of $\cvar$ can
be derived exactly, see \cite{cvarfordists} for formulas which we
repeat for completeness.  If $X$ follows a
Weibull$(\kappa,\lambda)$ distribution, then
\begin{equation*}
  \cvar(X) = \frac{\lambda}{1-\alpha} \Gamma\left(1+\frac{1}{\kappa}, -\log (1-\alpha)\right),
\end{equation*}
where $\Gamma(a,b) = \int_{b}^{\infty} p^{a-1} e^{-p} \,dp$ is the
upper incomplete gamma function Moreover, if $X$ follows a
lognormal$(\mu,\sigma)$ distribution, then
\begin{equation*}
  \cvar(X) = \frac{e^{\mu + \sigma^2/2}}{1-\alpha} \Phi \left[ \sigma - \frac{\Phi^{-1}(\alpha)}{\sqrt{2}}\right],
\end{equation*}
where $\Phi$ and $\Phi^{-1}$ are respectively the standard normal CDF
and its inverse.

For the GPD distribution, the tail distribution is exactly GPD
distributed as explained in \cref{le:distrGPinfo}, and therefore the
EVT approximation of the $\cvar$ is asymptotically unbiased (i.e. as
the number of stages tends to infinity). For the Weibull and lognormal
distributions, the EVT approximation is clearly biased, and the
simulation experiments shall help investigating whether the reduction
in variance provided by the EVT in comparison to the SA method is
sufficient to offset the bias of the former method.

The performance of estimates is assessed using two metrics. For
$m=1,\ldots,M$, denote the stage-$t$ estimate of the arm $j$ $\cvar$
for run $m$ by ${}_m\widehat{\cvar^t}(X_j)$. The first is the commonly
used root-mean-square error (RMSE):
$$\text{RMSE}_t = \sqrt{ \frac{1}{M} \sum_{m=1}^{M} \left({}_m\widehat{\cvar^t}(X_j) - \cvar(X_j)\right)^2}$$
Since the RMSE is sensitive to outliers, a second metric is also
considered: the percentage of times that the EVT $\cvar$ estimate is
closer to the true value of the $\cvar$ than the SA estimate across all
runs. We refer to this metric as \textit{Fraction Closer}
subsequently.

To summarize the simulation procedure, for each run $m$, at each stage
$t$, calculations are performed on the first $t$ observations
$\{x_1, \ldots, x_t\}$ with the following procedure:
\begin{enumerate}
\item For the EVT estimate, consider a set of candidate thresholds
  $u_1, \ldots, u_l$.
\item For each possible value of $u$, calculate the threshold excesses
  $x_i - u$, $i = 1,...,t$ and use the MLE to estimate parameters for
  GPD of excesses. This leads to the selection of the optimal
  threshold $u$ through the method described in \cref{se:threshold}.
\item Calculate $\cvar$ estimates using the SA and EVT methods.
\end{enumerate}

The confidence level of the CVaR in the simulation experiments is set
to $\alpha=0.999$.
A high confidence level is considered since the scarcity of
observations is more important for such levels; this is where the EVT
method is most likely to outperform the SA counterpart and prove the
most useful.  In all simulations, at stage $t$, $u_1$ and $u_l$ are
respectively set to the $0.7$ and $\alpha$ confidence level sample quantiles of
the empirical distribution of costs sampled previously in the run from
the arm. The number of threshold considered is set $l=50$, and the
threshold $u_j$ is set as the empirical cost distribution quantile
with confidence level
$\tilde\alpha_j = \tilde\alpha_1 + (\tilde\alpha_l-\tilde\alpha_1)
\frac{j-1}{l}$, $j=1,\ldots l$; equally spaced threshold confidence
levels spanning the interval $[0.7,\alpha]$ are used.  The ForwardStop
rule confidence level $\gamma$ was set to 0.1.

To provide additional stability to the EVT approach, a small
modification to the threshold procedure was applied.  Whenever for a
given candidate threshold $u$ the maximum likelihood estimates
\eqref{eq:MLestim} for exceedances are such that $\hat{\xi} > 0.9$,
the threshold $u$ was automatically discarded.  This is due to the
expression $1-\xi$ found at the denominator of the CVaR approximation
\ref{eq:TCEapproxEVT} which can make the estimate explode when
$\hat{\xi}$ is close to one. Although this comes at the expense of
generating some additional bias when the $\xi$ associated with the
limiting distribution is greater than $0.9$, this modification to the
algorithm never reduced its performance in some unreported tests
performed by the authors.

\begin{figure}[ht!]
\centering
  \begin{subfigure}{0.35\textwidth}
    \includegraphics[width=\textwidth]{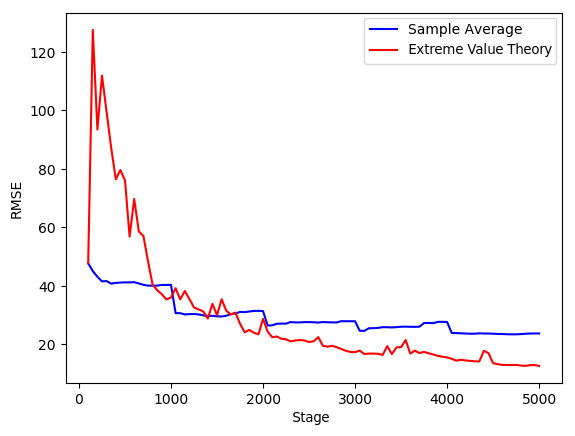}
    \caption{$\xi=0.4, \sigma=1$}
    \label{fig:gpd1}
  \end{subfigure}
  \begin{subfigure}{0.35\textwidth}
    \includegraphics[width=\textwidth]{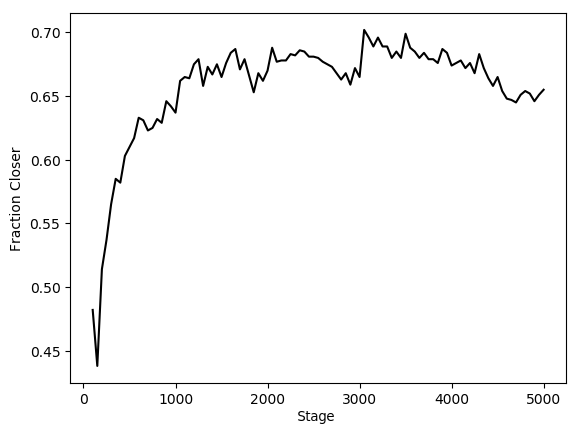}
    \caption{$\xi=0.4, \sigma=1$}
    \label{fig:gpd2}
  \end{subfigure}
  \begin{subfigure}{0.35\textwidth}
    \includegraphics[width=\textwidth]{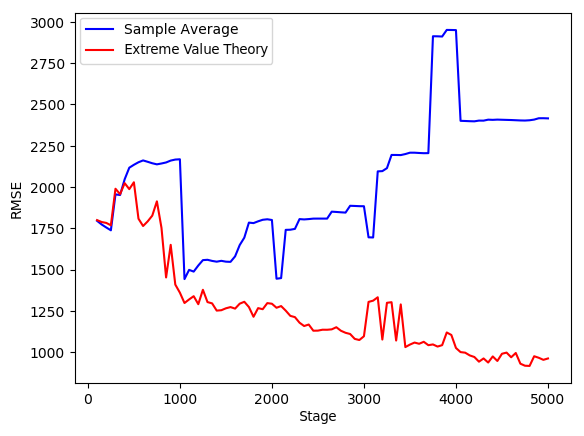}
    \caption{$\xi=0.8, \sigma=1$}
    \label{fig:gpd3}
  \end{subfigure}
  \begin{subfigure}{0.35\textwidth}
    \includegraphics[width=\textwidth]{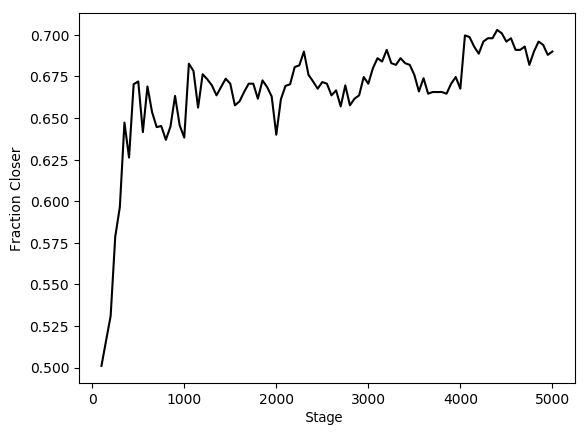}
    \caption{$\xi=0.8, \sigma=1$}
    \label{fig:gpd4}
  \end{subfigure}
  \caption{RMSE and Fraction Closer at each stage in the single-arm
    simulation experiment for the generalized Pareto distribution with
    parameters $\xi$ and $\sigma$.}\label{fig:gpd}
\end{figure}

\begin{figure}[ht!]
\centering
  \begin{subfigure}{0.35\textwidth}
    \includegraphics[width=\textwidth]{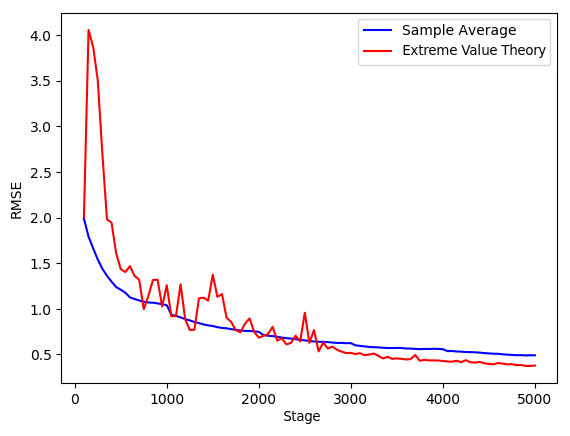}
    \caption{$\mu=0, \sigma=0.5$}
    \label{fig:lnorm1}
  \end{subfigure}
  \begin{subfigure}{0.35\textwidth}
    \includegraphics[width=\textwidth]{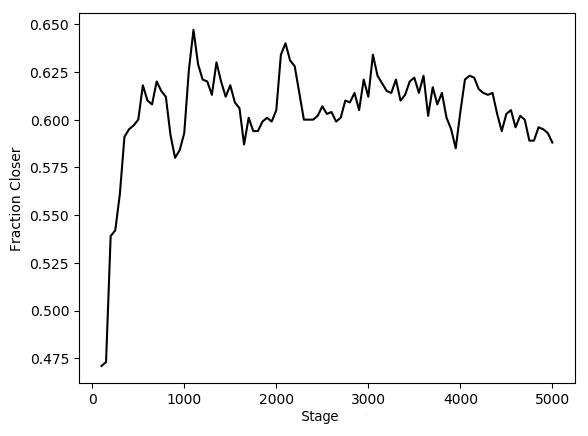}
    \caption{$\mu=0, \sigma=0.5$}
    \label{fig:lnorm2}
  \end{subfigure}
  \begin{subfigure}{0.35\textwidth}
    \includegraphics[width=\textwidth]{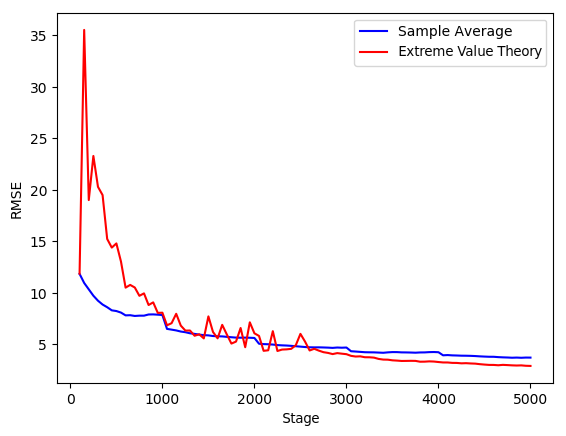}
    \caption{$\mu=0, \sigma=0.9$}
    \label{fig:lnorm3}
  \end{subfigure}
  \begin{subfigure}{0.35\textwidth}
    \includegraphics[width=\textwidth]{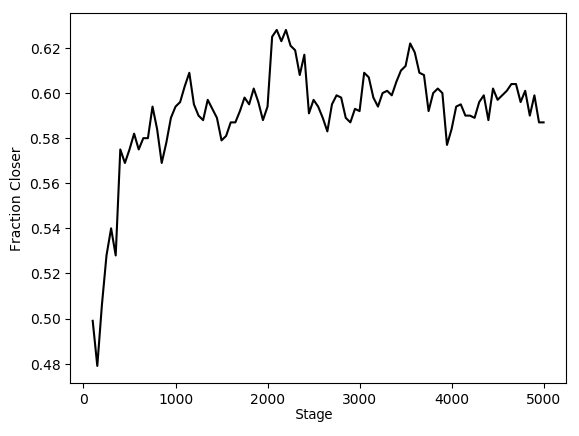}
    \caption{$\mu=0, \sigma=0.9$}
    \label{fig:lnorm4}
  \end{subfigure}
  \caption{RMSE and Fraction Closer at each stage in the single-arm
    simulation experiment for the lognormal distribution with
    parameters $\mu$ and $\sigma$.}\label{fig:lnorm}
\end{figure}

\begin{figure}[ht!]
\centering
  \begin{subfigure}{0.35\textwidth}
    \includegraphics[width=\textwidth]{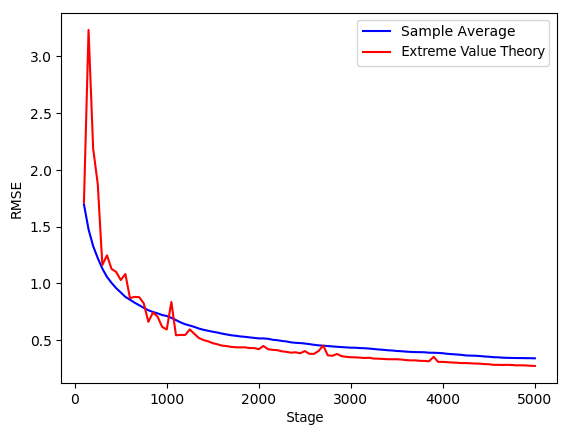}
    \caption{$\kappa=1.25, \lambda=1$}
    \label{fig:weib1}
  \end{subfigure}
  \begin{subfigure}{0.35\textwidth}
    \includegraphics[width=\textwidth]{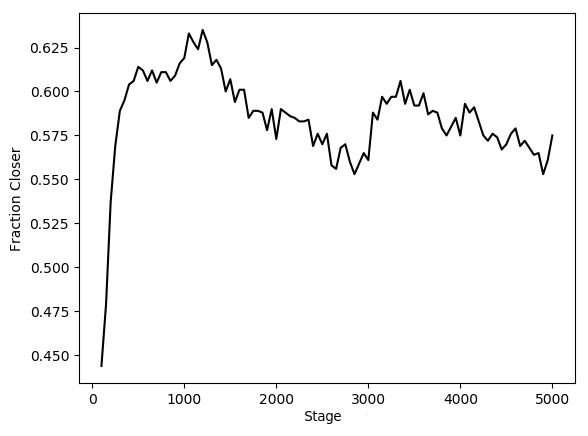}
    \caption{$\kappa=1.25, \lambda=1$}
    \label{fig:weib2}
  \end{subfigure}
  \begin{subfigure}{0.35\textwidth}
    \includegraphics[width=\textwidth]{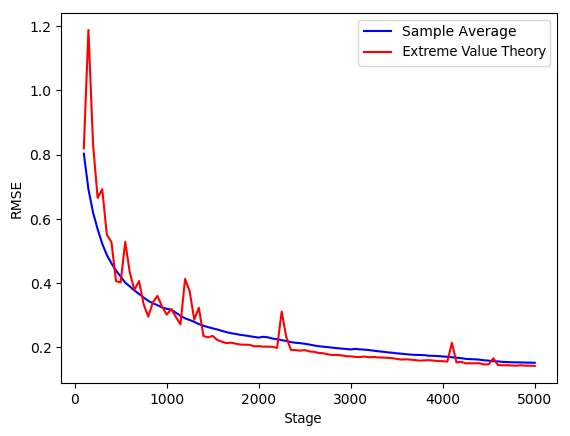}
    \caption{$\kappa=1.75, \lambda=1$}
    \label{fig:weib3}
  \end{subfigure}
  \begin{subfigure}{0.35\textwidth}
    \includegraphics[width=\textwidth]{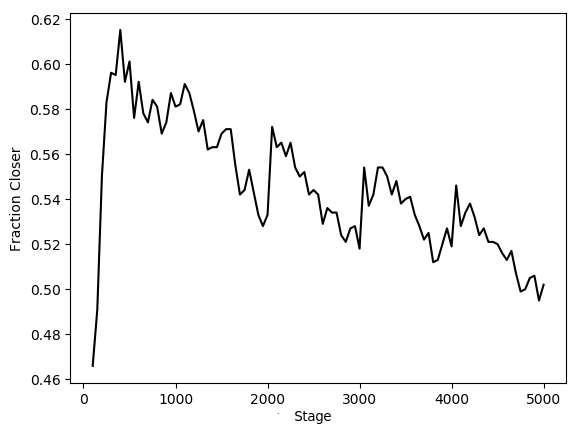}
    \caption{$\kappa=1.75, \lambda=1$}
    \label{fig:weib4}
  \end{subfigure}
  \caption{RMSE and Fraction Closer at each stage in the single-arm
    simulation experiment for the Weibull distribution with shape
    parameter $\kappa$ and scale parameter
    $\lambda$.}\label{fig:weibull}
\end{figure}

Figures \ref{fig:gpd}-\ref{fig:weibull} show results of running the
simulation study with various parameter configurations for the GPD,
lognormal, and Weibull distributions respectively.

\begin{figure}[ht!]
  \centering

  \begin{subfigure}{0.45\textwidth}
    \includegraphics[width=\textwidth]{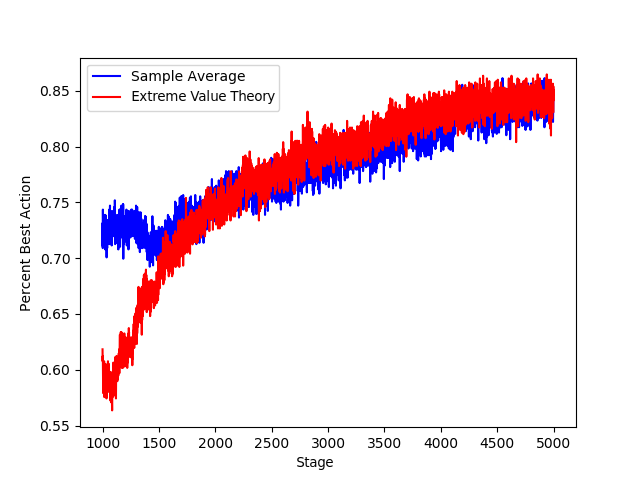}
    \caption{Underlying arm distributions: Lognormal with $\mu$=1 and
      $\sigma \in \{0.5, 0.6, 0.7, 0.8, 0.9\}$}
    \label{fig:lnorm_bandit}
  \end{subfigure}
  \begin{subfigure}{0.45\textwidth}
    \includegraphics[width=\textwidth]{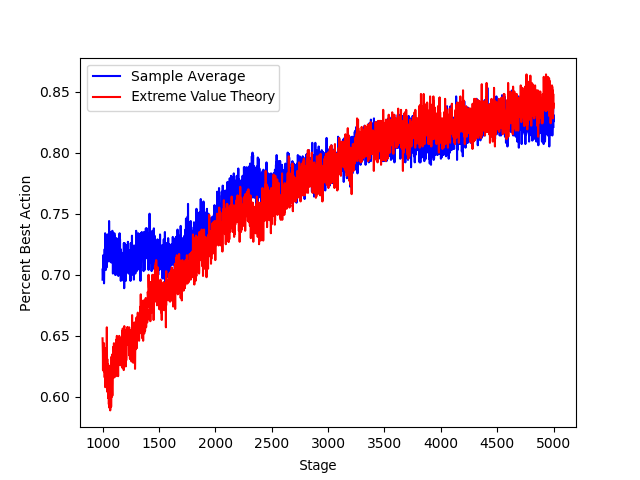}
    \caption{Underlying arm distributions: Weibull with $\lambda$=1
      and $\kappa \in \{0.75, 1.0, 1.25, 1.5, 1.75\}$}
    \label{fig:weibull_bandit}
  \end{subfigure}
   \begin{subfigure}{0.45\textwidth}
    \includegraphics[width=\textwidth]{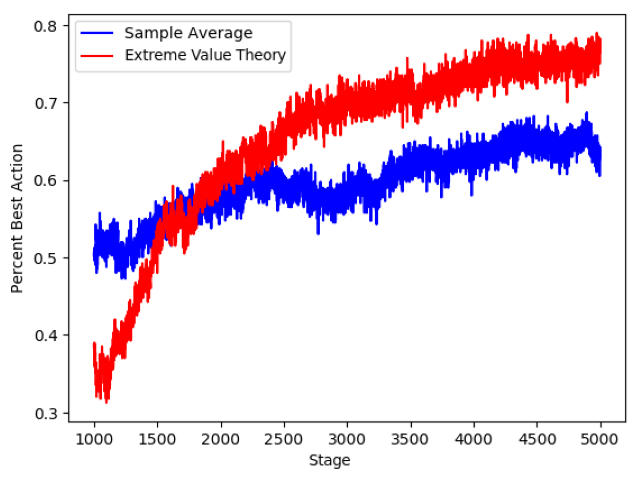}
    \caption{Underlying arm distributions: GPD with $\sigma$=1 and
      $\xi \in \{0.4, 0.5, 0.6, 0.7, 0.8\}$}
    \label{fig:gpd_bandit}
  \end{subfigure}
  \caption{Percent Best Action for both the Sample Average (SA)
    and Extreme Value Theory (EVT) CVaR estimation methods in three
    5-arm MAB simulations, where the underlying arm distributions are respectively lognormal, Weibull and GPD.}\label{fig:bandits}
\end{figure}

A general observation which can be made is that for most of the tested
parameter configurations, the EVT method
tends to under-perform and exhibit less stability in
earlier stages in terms of RMSE compared to SA. However, at subsequent stages, the
EVT estimate tends to stabilize and eventually provides better
performance than the SA estimate. The same phenomenon is observed when
looking at the Fraction Closer metric. An interesting observation is
that EVT starts outperforming the SA according to the Fraction Closer
earlier than it does in terms RMSE. Since the RMSE is very sensitive
to large errors contrarily to the Fraction Closer, this tends to
indicate that the EVT approach can lead to larger errors than the SA
before it stabilizes. This could partly be due to a large EVT
estimator variance in early stages when the estimate $\hat{\xi}$ is
not very precise and can take values close to the $0.9$ limit that was
set; this would lead to very large CVaR estimates due to the
reciprocal of $1-\xi$ found in \eqref{eq:TCEapproxEVT} as mentioned
previously.
 
\subsection{Best-arm selection in a multi-arm bandit simulation}

In the current section, results from a $5$-arm MAB simulation are provided. This experiment is analogous to the one from
\cref{se:singlearm}, except there are now $k=5$ arms from which to
sample costs instead of one. The cost distribution is different for
each arm, and thus a distinct estimate for the CVaR is formed for each
of the arms. The arm selection policy considered is the
$\epsilon$-greedy one described in \cref{sec:MultiArmBandits}. To
encourage exploration, a fully random arm selection is used for the
first 1000 stages, whereas for subsequent stages the exploration
probability is set to $0.1$. This corresponds to the
schedule
\begin{equation*}
  \epsilon_t = \begin{cases}
    1, \quad t=1,\ldots, 1000,
    \\ 0.1, \quad t=1001,\ldots, 5000.
  \end{cases}
\end{equation*}

Again, three experiments are performed, where arm cost
distributions are respectively GPD, lognormal or Weibull. For the
GPD, $\sigma=1$ is kept fixed across all arms, while the tail varies
across arms, taking values $\xi = 0.4, 0.5, 0.6, 0.7, 0.8$.  For the
lognormal distribution, the location parameter $\mu=1$ is kept fixed
whereas the scale parameter takes respective values
$\sigma = 0.5, 0.6, 0.7, 0.8, 0.9$ across arms. Finally, for the
Weibull distribution, $\lambda=1$ is set for all arms whereas
$\kappa =0.75, 1.0, 1.25, 1.5, 1.75$ varies across the five arms.

The performance metric considered for the MAB experiments
is referred to as the \textit{Percent Best Action}, which represents
the percentage of time across all runs that the least risky arm is selected
at a given stage $t$. This is a useful metric since it provides an estimate of the probability of selecting the optimal arm after $t$ time steps. Figure \ref{fig:bandits} provides values
obtained for that metric for each of the three experiments at all
stages of the simulation.

The main lesson obtained from the multi-arm bandit simulation
results is qualitatively the same as for the single-arm experiment:
for early stages, the SA method performs better than the EVT, but the
EVT eventually catches up and outperforms the SA in its ability to
select the less risky arm. This clearly demonstrates the usefulness of
considering an EVT estimation method for the CVaR when considering a
multi-arm bandit action selection framework.

\section{Conclusion}\label{se:conclusion}
We have investigated the use of a CVaR estimator based on extreme value theory in a risk-averse multi-arm bandit problem. Using the generalized Pareto approximation of a distribution's tail, we established a new estimation procedure for the CVaR which we call the EVT CVaR. While the derivation of the CVaR approximation in \cref{eq:TCEapproxEVT} exists in the literature, its efficacy in statistical estimation is limited by the problem of threshold selection, which can be unreliable in practice. The novelty of our approach from a computational perspective is to integrate the sequential goodness-of-fit test of \cite{bader2018automated} in CVaR estimation using the GPD approximation. We have shown empirically that the EVT CVaR leads to reliable estimates and performance improvements compared to the more commonly encountered sample average CVaR estimator in some distributions. In the MAB setting, we showed using a simple $\epsilon$-greedy policy that the EVT CVaR can also be a preferable choice for action selection under risk criteria when the CVaR confidence level $\alpha$ is very high.

\section*{Acknowledgements}
Financial support from NSERC (Godin, RGPIN-2017-06837; Yu, RGPIN-2018-05096) is gratefully acknowledged. We would like to thank Debbie J. Dupuis for her extremely valuable feedback.
\bibliographystyle{apalike} \bibliography{Bib}

\appendix

\section{Confidence intervals for the CVaR estimates}

\subsection{Sample CVaR confidence interval}
\label{se:STCECI}

The bootstrapping procedure for the construction of a confidence
interval around the sample CVaR estimate entails resampling $M$
samples with replacement of size $t$ from $S_t$, with $M$ being a
large integer. Denoting the $m^{th}$ bootstrapped sample by
$S^{(m)}_t \equiv \{ y^{(m)}_1,\ldots,y^{(m)}_t \}$ with
$m=1,\ldots,M$, a CVaR estimate can be obtained for each new sample:
\begin{eqnarray*}
  \widehat{\cvar}^{(m)} &=& \frac{\sum_{i=1}^{t} y^{(m)}_{i} \mathds{1}_{ \{y^{(m)}_i \geq \hat{q}^{(m)}_{\alpha} \} }}{\sum_{i=1}^{t} \mathds{1}_{ \{y^{(m)}_i \geq \hat{q}^{(m)}_{\alpha} \} }}, \text{ where } 
                           \hat{q}^{(m)}_{\alpha} \equiv   y^{(m)}_{(\lceil\alpha t\rceil)}
\end{eqnarray*}
with $y^{(m)}_{(1)}, \ldots, y^{(m)}_{(t)}$ are the respective order
statistics of $S^{(m)}_t$. Denote $v_1,\ldots,v_M$ the order
statistics of the set $\big\{ \widehat{\cvar}^{(m)} \big\}^M_{m=1}$.
Then, a bilateral confidence band of confidence level $\tilde{\alpha}$
for $\cvar(Y)$ is given by
$\left[v_{(\lceil M(1-\tilde{\alpha})/2 \rceil)} , v_{(\lceil
    M\tilde{\alpha}/2 \rceil)} \right]$.

\subsection{Extreme Value Theory CVaR confidence interval}
\label{se:EVTCI}

Assuming the exactness of the approximation of the tail distribution
by a GPD (i.e. ignoring the misspecification), the maximum likelihood
estimates $(\hat{\xi},\hat{\sigma})$ from \eqref{eq:MLestim} have the
following asymptotically behavior:
\begin{equation*}
  \sqrt{N_u} \left( [\hat{\xi},\hat{\sigma}]^\top - [\xi,\sigma]^\top\right) \Rightarrow N(0,\mathcal{I}^{-1})
\end{equation*}
as $N_u \rightarrow \infty$, where $\Rightarrow$ denotes convergence
in law, $N$ is the Gaussian distribution and $\mathcal{I}^{-1}$ is the
inverse of the Fisher information matrix
\begin{equation*}
  \mathcal{I} \equiv -\mathbb{E}\left[
    \begin{array}{cc}
      \frac{\partial^2 }{\partial \xi^2} \log g_{\xi,\sigma} (Z)&  \frac{\partial^2 }{\partial \xi \partial \sigma} \log g_{\xi,\sigma} (Z) \\
      \frac{\partial^2 }{\partial \xi \partial \sigma} \log g_{\xi,\sigma} (Z) &  \frac{\partial^2 }{\partial \sigma^2} \log g_{\xi,\sigma} (Z) 
    \end{array}
  \right]
  \approx -\frac{1}{N_u} \sum_{j=1}^{N_u} \left[
    \begin{array}{cc}
      \frac{\partial^2 }{\partial \xi^2} \log g_{\xi,\sigma} (z_{j,u})&  \frac{\partial^2 }{\partial \xi \partial \sigma} \log g_{\xi,\sigma} (z_{j,u}) \\
      \frac{\partial^2 }{\partial \xi \partial \sigma} \log g_{\xi,\sigma} (z_{j,u}) &  \frac{\partial^2 }{\partial \sigma^2} \log g_{\xi,\sigma} (z_{j,u}) 
    \end{array}
  \right]
\end{equation*}
where $Z$ is a random variable whose distribution is a
GPD$(\xi,\sigma)$.

Partial derivatives from the information matrix can be developed as
follow for the cse $\xi \neq 0$:
\begin{eqnarray*}
  \frac{\partial }{\partial \sigma} \log g_{\xi,\sigma} (z) &=& \frac{1}{\sigma} \left[\frac{z(\xi+1)}{\sigma + \xi z}-1\right],
  \\ \frac{\partial }{\partial \xi} \log g_{\xi,\sigma} (z) &=&  \frac{1}{\xi^2} \log\left (1+\frac{\xi z}{\sigma}\right )
                                                                -\left(\frac{1}{\xi} + 1\right)  \frac{z}{\sigma+\xi z},
  \\ \frac{\partial^2 }{\partial \sigma^2} \log g_{\xi,\sigma} &=& -\frac{1}{\sigma^2} \left[\frac{z(\xi+1)}{\sigma + \xi z}-1\right] - \left[\frac{z(\xi+1)}{\sigma(\sigma + \xi z)^2}\right],
  \\ \frac{\partial^2 }{\partial \sigma \partial \xi} \log g_{\xi,\sigma} &=& \frac{1}{\sigma} \left[\frac{z}{\sigma + \xi z} - \frac{z^2(\xi+1)}{(\sigma + \xi z)^2}\right],
  \\ \frac{\partial^2 }{\partial \xi^2} \log g_{\xi,\sigma} &=& -\frac{2}{\xi^3} \log\left (1+\frac{\xi z}{\sigma}\right ) 
                                                                +\frac{1}{\xi^2} \frac{z}{\sigma+\xi z}
                                                                +  \frac{z}{\xi^2(\sigma+\xi z)}
                                                                +\left(\frac{1}{\xi} + 1\right)  \frac{z^2}{(\sigma+\xi z)^2}.
\end{eqnarray*}

From the delta-method \citep[see for instance Appendix B.3.4.1 in
][]{remillard2016statistical}, for a well-behaved function
$h: \mathbb{R}\times (0,\infty )\rightarrow \mathbb{R}$,
\begin{equation*}
  \sqrt{N_u} \left( h(\hat{\xi},\hat{\sigma}) - h(\xi,\sigma) \right) \Rightarrow N \left(0, [\nabla h(\xi,\sigma)]^\top \mathcal{I}^{-1} [\nabla h(\xi,\sigma)]\right)
\end{equation*}
where $[\nabla h(\xi,\sigma)]$ is the column vector representing the
gradient of $h$.

Setting
\begin{equation}
  h(\xi,\sigma) \equiv q + \frac{\sigma + \xi(q-u)}{1-\xi}
\end{equation}
as in \eqref{eq:TCEapproxEVT} yields
\begin{equation*}
  \frac{\partial }{\partial \xi}h(\xi,\sigma) = \frac{q-u+\sigma}{(1-\xi)^2}, \qquad \frac{\partial }{\partial \sigma}h(\xi,\sigma) = \frac{1}{1-\xi}.
\end{equation*}

Combining all previous results and disregarding the variability of
$\hat{q}_\alpha$ implies that
\begin{equation*}
  Var[h(\hat{\xi},\hat{\sigma})] \approx \frac{1}{N_u} [\nabla h(\hat{\xi},\hat{\sigma})]^\top \mathcal{I}^{-1} [\nabla h(\hat{\xi},\hat{\sigma})]
\end{equation*}
which can be used to obtain a Gaussian asymptotic confidence interval
for $\cvar(Y)$.

\section{Proofs}
\label{se:Proofs}

The following Lemma \cref{le:distrGPinfo} can then be used to obtain the CVaR of a
Generalized Pareto distribution.

\begin{lemma}[see \citealp{mcneil2005quantitative}]
  \label{le:distrGPinfo}
  Let $Y$ be random variable with a Generalized Pareto distribution
  with parameters $(\xi, \sigma)$, i.e. $F_Y(y) = G_{\xi, \sigma}(y)$,
  where the latter CDF is defined in \eqref{def:GP}. Then,
  \begin{equation*}
    \mathbb{E}[Y] = \frac{\sigma}{1-\xi} \quad \text{ if } \xi<1.
  \end{equation*}
  Moreover, consider any $u \in [0,\infty)$ if $\xi \geq 0$ or any
  $u \in [0,-\sigma / \xi]$ if $\xi<0$. Then the conditional
  distribution of $Y-u$ given $Y>u$ is a Generalized Pareto
  distribution with parameters $(\xi, \sigma+\xi u)$, i.e.,
  \begin{equation*}
    1-K_{u}(y) = \frac{1-G_{\xi, \sigma}(y+u)}{1-G_{\xi, \sigma}(u)} = 1-G_{\xi, \sigma+ \xi u}(y).
  \end{equation*}
\end{lemma}

\begin{corollary}
  \label{co:TCEGPder}
  Assume $F_Y(y) = G_{\xi, \sigma}(y)$ with $\xi<1$ and
  $\sigma>0$. Consider $u>0$ such that $\sigma+\xi u >0$. Then,
  \begin{equation*}
    \mathbb{E}[Y \vert Y > u] = u + \frac{\sigma + \xi u}{1-\xi}
  \end{equation*}
\end{corollary}

\textbf{Proof of \cref{cor:approxTCE}}: First,
\begin{eqnarray*}
  \cvar(Y) = \mathbb{E}[Y | Y \geq q_{\alpha}]
  = u + \mathbb{E}[Y -u| Y \geq q_{\alpha}] = u + \mathbb{E}[Y -u| Y-u \geq q_{\alpha}-u].
\end{eqnarray*}
Since $q_{\alpha} \geq u$, $Y-u \geq q_{\alpha}-u$ implies that
$Y \geq u$. Furthermore, the CDF of $Y-u$ given $Y \geq u$ is
approximately $G_{\xi,\beta(u)}$ for some mapping $\beta$ by
\cref{PickhandsBalkema}.

Therefore defining a random variable $Z$ having the CDF
$G_{\xi,\beta(u)}$ (i.e. approximating the distribution of the
exceedance $Y-u$),
\begin{align*}
  \cvar(Y) &\approx u + \mathbb{E}[Z| Z \geq q_{\alpha}-u]\\
  (\textrm{by \cref{co:TCEGPder}}) &= u + (q_{\alpha}-u) + \frac{\beta(u) + \xi (q_{\alpha}-u)}{1-\xi}\\
          &= q_{\alpha} + \frac{\beta(u) + \xi (q_{\alpha}-u)}{1-\xi}.
\end{align*}
$\square$

\textbf{Proof of \cref{co:VarApprox}}: First, from
\cref{PickhandsBalkema}, the distribution of $Y-u$ given $Y>u$ is
approximately GPD. Using this approximation, since $q_\alpha \geq u$
would have no atoms in a neighborhood around $q_\alpha$ and therefore
$\alpha \approx F_Y\left(q_{\alpha}\right)$. absolutely
continuous. This implies by conditioning that
\begin{align*}
  1-\alpha &\approx 1-F_Y\left(q_{\alpha}\right) 
  \\ &= \left(1-K_{u}\left(q_{\alpha}-u\right) \right) \left( 1-F_Y\left(u\right)\right) 
  \\ &\approx \left(1-G_{\hat{\xi}, \hat{\sigma}}\left(\hat{q}_{\alpha}-u\right) \right) \left( 1-\hat{F}^t_Y\left(u\right)\right)
\end{align*}
which implies
\begin{eqnarray*}
  G_{\hat{\xi}, \hat{\sigma}}\left(\hat{q}_{\alpha}-u\right) &\approx& 1-\frac{1-\alpha}{1-\hat{F}^t_Y\left(u\right)} 
  \\ \Rightarrow 1-\left(1+\frac{\hat{\xi}\left(\hat{q}_{\alpha}-u\right)}{\hat{\sigma}}\right)^{(-1 / \hat{\xi})} &\approx& \frac{\alpha-\hat{F}^t_Y\left(u\right)}{1-\hat{F}^t_Y\left(u\right)}.
\end{eqnarray*}
Isolating $\hat{q}_{\alpha}$ in the latter expression directly leads
to \eqref{eqcorVaR}.

$\square$

\textbf{Proof of the lognormal CVaR formula}:

Let $\text{erf}$ denote the error function which is related to the
standard normal CDF $\Phi$ through
\begin{equation*}
  \Phi(x) = \frac{1}{2} \left[1+ \text{erf} \left(\frac{x}{\sqrt{2}}\right)\right]
\end{equation*}
which implies
\begin{eqnarray}
  \label{erfphiequiv}
  \text{erf}(x) = 2 \Phi (\sqrt{2} x) -1, \quad \text{erf}^{-1}(x) = \frac{1}{2} \Phi^{-1} \left[ \frac{x+1}{2}\right].
\end{eqnarray}
If $X$ follows a lognormal$(\mu,\sigma)$ distribution,
\cite{cvarfordists} show in their Proposition 9 that the CVaR of $X$
is given by
\begin{equation*}
  \cvar(X) = \frac{e^{\mu + \sigma^2/2}}{2(1-\alpha)} \left[1+ \text{erf} \left( \frac{\sigma}{\sqrt{2}} - \text{erf}^{-1} (2 \alpha-1) \right)\right].
\end{equation*}
which, using \eqref{erfphiequiv}, leads to
\begin{equation*}
  \cvar(X) = \frac{e^{\mu + \sigma^2/2}}{1-\alpha} \Phi \left[ \sigma - \frac{\Phi^{-1}(\alpha)}{\sqrt{2}}\right].
\end{equation*}

$\square$

\end{document}